\documentclass[
]{ceurart}

\usepackage{listings}
\usepackage{mathtools}
\usepackage{todonotes}
\usepackage{hyperref}
\usepackage{cleveref}
\usepackage{csquotes}
\usepackage{caption}
\usepackage{subcaption}

\begin{document}

\copyrightyear{2022}
\copyrightclause{Copyright for this paper by its authors.
  Use permitted under Creative Commons License Attribution 4.0
  International (CC BY 4.0).}

\conference{In A. Martin, K. Hinkelmann, H.-G. Fill, A. Gerber, D. Lenat, R. Stolle, F. van Harmelen (Eds.), 
Proceedings of the AAAI 2022 Spring Symposium on Machine Learning and Knowledge Engineering for Hybrid Intelligence (AAAI-MAKE 2022), 
Stanford University, Palo Alto, California, USA, March 21–23, 2022.}

%\title{Multimodal analysis and understanding of human-object interactions by an autonomous agent: detecting and fulfilling user needs.}

\title{Commonsense Reasoning for Identifying and Understanding the Implicit Need of Help and Synthesizing Assistive Actions}

\author[1,2,3]{Maëlic Neau}[%
email=neau@enib.fr,
]

\author[2]{Paulo Santos}[email=paulo.santos@flinders.edu.au,url=https://www.flinders.edu.au/people/paulo.santos,]

\author[1]{Anne-Gwenn Bosser}[email=anne-gwenn.bosser@enib.fr,]

\author[3]{Nathan Beu}[email=nathan.beu@adelaide.edu.au,]

\author[1,3]{Cédric Buche}[email=buche@enib.fr,url=https://www.enib.fr/~buche/]

\address[1]{Lab-STICC/ENIB, France}
\address[2]{College of Science and Engineering, Flinders University of South Australia, \\1284 South Rd, Clovelly Park SA 5042, Australia}
\address[3]{CNRS, International Research Lab "CROSSING", Adelaide, Australia}

\begin{abstract}
Human-Robot Interaction (HRI) is an emerging subfield of service robotics. While most existing approaches rely on \textit{explicit} signals (i.e. voice, gesture) to engage, current literature is lacking solutions to address \textit{implicit} user needs. In this paper, we present an architecture to (a) detect user implicit need of help and (b) generate a set of assistive actions without prior learning. Task (a) will be performed using state-of-the-art solutions for Scene Graph Generation coupled to the use of commonsense knowledge; whereas, task (b) will be performed using additional commonsense knowledge as well as a sentiment analysis on graph structure. Finally, we propose an evaluation of our solution using established benchmarks (e.g. ActionGenome dataset) along with human experiments. The main motivation of our approach is the embedding of the perception-decision-action loop in a single architecture.
\end{abstract}


\maketitle

\section{Introduction} \label{intro} %% 1 page 

%To come up with a system able to detect the implicit need of help, a solution could be deep learning (supervised, unsupervised or self-supervised). For instance, given a set of labeled or unlabeled data where one human is performing a task and another one is helping him, the system should learn the relevant parameters to the concept "need of help" as well as "providing help". Such a solution comes with a main drawback: data collection. 
%Because we do not want to restrict our perspective to a specific sort of task being performed, the creation of a large and diverse dataset would be arduous. Data augmentation \cite{shorten2019survey} or few-shot learning \cite{ravi2016optimization} are two approaches that could reduce this gap. Still we believe that traditional deep learning is not the best approach for our proposal as the real world is much more changing and unpredictable than one could retrieves in scientific experiments.

Detecting and understanding user's intentions and needs is the fundamental backbone of service robotics. This question relates to how  high-level, abstract, concepts can be inferred from raw sensor data (an issue intimately related to the symbol grounding problem) \cite{harnad1990symbol}.
Traditional approaches to this problem in robotics use \textit{explicit} signals from the user such as voice \cite{tada2020robust}, gesture \cite{waldherr2000gesture} or even touch \cite{doisy2012sensorless}. However, the deployment of service robots in assisting activities of daily life (ADL), especially for impaired or elderly people, is leading the way to more \textit{implicit} interactions with autonomous agents.
Previous work has been introduced to understand user's implicit intentions in service robotics. Some use external context to predict user's intentions \cite{Liu_Zhang_Li_2014} while others rely on gaze-based signals \cite{Li_Zhang_2017}. But to the best of our knowledge none of them integrates the use of external commonsense knowledge of the world.

The present paper tackles an important part of this issue, where only non-verbal, visually-observable data are taken into account. We present a system that is able to understand the implicit user needs of help in the realisation of a task and to provide a relevant assistive action, inspired by the way humans act based on commonsense reasoning. 
In a general sense, the use of commonsense reasoning in the present work can be summarised with the following assumption: a factor of human assistance to one another in the realisation of a task is the perception of danger. For instance,
humans are typically able to understand (without explicit prior learning) that anything coming out of an oven is hot, and that a person should protect their hands to avoid hurting themselves.  This is connected to the following definition of commonsense reasoning, from \cite{kuipers1984commonsense}, p.170:
\begin{quote}\em
Commonsense causal reasoning is qualitative reasoning about the behavior of a mechanism which can be done without external memory or calculation aids, although it may draw on concepts learned from the advanced study of a particular domain, e.g. automobile mechanics, computer architecture, or medical physiology.
\end{quote}
For instance, in the above example, we can summarise the commonsense reasoning as the following causal relationships:

\begin{equation}
  oven \xrightarrow[]{\text{produce}} heat
\end{equation}
\begin{equation}
  heat \xrightarrow[]{\text{capable of}} hurt \: skin
\end{equation}

With a refinement on the visual features, the system is able to ground commonsense knowledge to the scene as follow:

\begin{equation}
  heat \xrightarrow[]{\text{capable of}} hurt \: hand
\end{equation}

This commonsense reasoning process will also help the robotic agent to build an assistive action. In fact, in some cases, the  reasoning about the visual inputs alone is not sufficient to provide accurate help. Recalling the previous example, the system needs external knowledge to come up with the assistive action "bring gloves to the user", as a human would do:

\begin{equation}
  glove \xrightarrow[]{\text{capable of}} protect \: hand
\end{equation}

We believe that creating such relationships is possible by using commonsense knowledge databases such as the ConceptNet \cite{Speer_Chin_Havasi_2017} or ATOMIC \cite{Sap_Bras_Allaway_Bhagavatula_Lourie_Rashkin_Roof_Smith_Choi_2019} datasets, and also following some of the ideas for combining logic reasoning with machine learning described in \cite{van_Harmelen_Teije_2019}. Another characteristic of our work is the use of Scene Graph (SG) \cite{Johnson_Krishna_Stark_Li_Shamma_Bernstein_Fei-Fei_2015} as a tool for knowledge representation. In fact, this type of representation could easily be enriched with external knowledge databases as they share the same data structure: graph. Finally, the use of commonsense reasoning is the part of our architecture, which gives us the possibility of understanding biases and correcting them in an incremental development.

Our position could be summarized as the following statement: from the analysis of human behavior, the use of state-of-the-art solutions from Vision-to-Language combined with Commonsense Reasoning will leverage Cognitive Robotics.

\section{Related work} %% 2-3 pages

The task of retrieving graph representations from still images or videos is called Scene Graph Generation (SGG), in this section we review current approaches for SGG. To perform efficient reasoning, our solution integrates external commonsense knowledge. Thus, we also review solutions for knowledge-graph enrichment and completion. Finally, as our reasoning system needs to provide a sentiment analysis to retrieve the possibility of risks, we review approaches to connotation lexicon \cite{Feng_Bose_Choi_2011} (i.e. lexicon that lists words with connotative polarity).

\subsection{Scene Graph Generation}

Scene Graph Generation (SGG) \cite{Johnson_Krishna_Stark_Li_Shamma_Bernstein_Fei-Fei_2015} is the task of creating a grounded graph of visual entities retrieved from an image with the goal of representing attributes, objects and their relationships in a scene. Such graphs typically contain one or more triplet(s)  {\em (head entity, relation, tail entity)}. Entities present in a scene graph could be person (e.g. {\em woman}), place ({e.g. \em street}), object ({e.g. \em jeans}) or attributes ({e.g. \em blue, long}). Relations between entities are even spatial positions ({e.g. \em in front of, behind}), actions ({e.g. \em walking}) or descriptions ({e.g. \em wearing}). 

While recent approaches for this task may differ, the majority are using object detection and region captioning as baseline \cite{Li_Ouyang_Zhou_Wang_Wang_2017}. For object detection, the most reported solution is the use of pre-trained Faster-RCNN \cite{Ren_He_Girshick_Sun_2017}, a highly efficient Convolutional Neural Network (CNN) approach with Region Of Interest (ROI) pooling for object classification. Once objects are detected, the SGG solutions need to pair entities with one another and find the correct predicate to represent this relation. To do so, approaches such as Conditional Random Fields (CRF) \cite{Dai_Zhang_Lin_2017} or Transitionnal Embeddings (TransE) \cite{Zhang_Kyaw_Chang_Chua_2017} are used. Lately, Neural Networks have leveraged this task with new RNN/LSTM-based \cite{zellers2018neural} and Graph Convolution Networks (GCN) \cite{yang2018graph} approaches.

When working with videos, a complex structure is needed to model relationships between images. To this end, \cite{Wang_Wei_Li_Zhang_Huang_2020} use a Temporal Convolution Network (TCN) paired to a GCN for modeling within-image dependencies. In \cite{Teng_Wang_Li_Wu_2021} the authors use Target Adaptive Context Aggregation to relate entities to their spatio-temporal context.

\subsection{Commonsense Completion}

There are multiple ways to use external knowledge to enrich a scene graph. The task of Commonsense Completion is introduced in \cite{Li_Taheri_Tu_Gimpel_2016} to define the automatic completion of a knowledge graph using commonsense knowledge, in most cases retrieved from ConceptNet  \cite{Speer_Chin_Havasi_2017}. In this task the knowledge is directly added to the graph, creating new nodes and edges. In COMET \cite{Bosselut_Rashkin_Sap_Malaviya_Celikyilmaz_Choi_2019}, the authors describe a model that learns how to generate graph completion based on relationships between events from the ATOMIC \cite{Sap_Bras_Allaway_Bhagavatula_Lourie_Rashkin_Roof_Smith_Choi_2019} and ConceptNet \cite{Speer_Chin_Havasi_2017} datasets. This method uses a Transformer architecture, the model is trained with a dataset of graphs to predict the next node, given the input previous node and a relation \begin{math}R\end{math} from the set of relationships of ATOMIC. As the solution uses a Transformer architecture, the input and output are natural language sentences, with specific tokens to represent relations.
We can also see the task of Commonsense Completion of the SG as a Knowledge Graph Fusion. \cite{Zareian_Karaman_Chang_2020} propose a new approach to bridge knowledge and scene graphs using successive message passing on a Graph Neural Network (GNN).

%Approaches for this task use for instance Transitional Embedding (TransE) 
%Instead of completion, attempts have been made to integrate commonsense knowledge directly into a scene graph for more accurate relation generation \cite{zareian2020learning}. In \cite{Wan_Liang_Du_Liu_Ou_Wang_Pan_Zeng_2021} the authors modify the scene graph with commonsense knowledge using iterative attention mechanism. Note that the commonsense knowledge is not specifically added to the existing graph but used in the process to refine attributes or predicates. Instead of adding new nodes and edges, commonsense knowledge is used to modify existing relationships to give a more detailed prediction, i.e. "person on top of horse" will become "person riding horse".

\subsection{Word Connotations Lexicon}

\cite{Feng_Bose_Choi_2011} is the first attempt to build a connotation lexicon --- a lexicon that maps words and their intrinsic connotation. The proposed approach learns word connotation using connotative predicates, i.e. predicates that ensure that words often encountered with ones negatively connoted will also be negatively connoted. With this method, the algorithm only needs a small seed of labelled words and a database of texts to learn words' connotations. In \cite{Feng_Kang_Kuznetsova_Choi_2013} the authors extend this method using induction algorithms based on graph structures. The use of Random Walk based on HITS/PageRank, Label/Graph Propagation and Constraint Optimization is reported. With this approach, \cite{Feng_Kang_Kuznetsova_Choi_2013} propose to capture fine-grained inductions, reducing biases from the previous solution (e.g. the world "cure" is often associated with "disease" while not being negatively connoted).

\section{Detecting and fulfilling the implicit need of help} %% 2-3 pages

From an input video sequence, the system builds its own representation of the task being performed using state-of-the-art approaches of Vision-to-Language. Then, it reasons on this representation using commonsense knowledge to assess risks for a human: if the risks are too high, an assistive action is performed accordingly. 
This work aims to solve the following questions: (a) Scene Graph enrichment with related Commonsense Knowledge; (b) Scene Graph refinement upon visual features; (c) Sentiment Analysis from Scene Graph; (d) Action Generation from Commonsense Scene Graph.

\subsection{Scene Understanding}

\begin{figure}
\centering
  \includegraphics[width=0.9\linewidth]{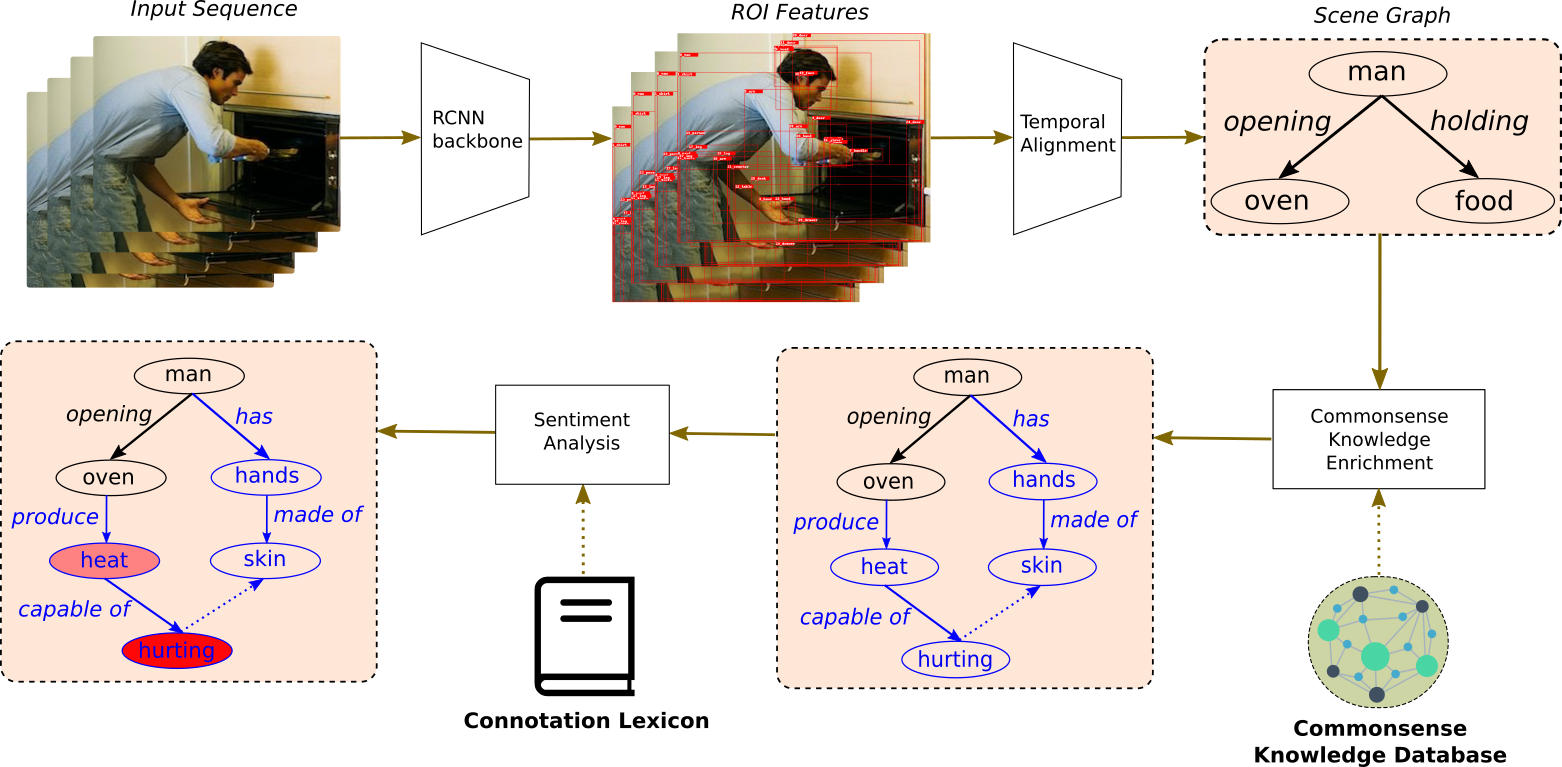}
  \caption{Our proposed architecture for scene representation and sentiment analysis.}
  \label{fig:sub1}
\end{figure}

The efficient scene understanding task from visual inputs, including human intents and objects affordances, is an old an unsolved challenge for Computer Vision. 
%The aim of this work is to provide a partial solution to this issue, where the use of commonsense knowledge inside the paradigm of service robotics. 
%As stated in other works \cite{Roy_Posner_Barfoot_Beaudoin_Bengio_Bohg_Brock_Depatie_Fox_Koditschek}, cognition and reasoning in robotics should not rely on traditional machine learning, instead build its own knowledge of the world, what we could call \textit{robot learning} \cite{Roy_Posner_Barfoot_Beaudoin_Bengio_Bohg_Brock_Depatie_Fox_Koditschek}. 
In this task, the use of an appropriate representation of the scene is of utmost importance. In the literature, there are two main approaches for scene representation: graphs and natural language processing; the former has been developed within Scene Graph Generation and the latter within Video Captioning. We choose scene graphs over natural language captions to model our representation of the scene. Scene Graphs provide numerous advantages: each detected entity is clearly represented and grounded, relationships between user and objects can be clearly identified and finally the enrichment of external knowledge is simple as most of knowledge bases also use graph structures \cite{Speer_Chin_Havasi_2017} \cite{Sap_Bras_Allaway_Bhagavatula_Lourie_Rashkin_Roof_Smith_Choi_2019}. 
In this section, we will detail our system for understanding the scene and inferring human's risks using SG, breaking down the description in four distinct steps (illustrated in Figure \ref{fig:sub1}):

\textbf{Scene Graph Generation} First, the representation of the relevant perceived data is critical. As a backbone, we use Faster-RCNN \cite{Ren_He_Girshick_Sun_2017} (Figure \ref{fig:sub1} top left) to retrieve ROI features from the input scene. Given these features, we need to construct a directed graph \begin{math} G \end{math} (Figure \ref{fig:sub1} top right) composed of a set of entities \begin{math} E \end{math} and a set of relations \begin{math} R \end{math} such that:

\begin{equation}
  G = (R, E, \theta)
\end{equation}

where \begin{math} \theta \end{math} is an incidence function that finds the relation \begin{math} r \in R \end{math} between the head entity \begin{math} h \in E \end{math} and the tail entity \begin{math} t \in E \end{math} such as:
\begin{equation}
 \theta : \{h,t\} \in E^2 \rightarrow{} r
\end{equation}
%In Scene Graph Generation (SGG), and also in more general knowledge graph-related tasks, a popular incidence function is the Transitional Embedding (TransE).
%TransE uses embedding into low-dimensional vector spaces to map relations between entities. 
To generate such a graph, we follow the approach from \cite{Teng_Wang_Li_Wu_2021} that uses Target Adaptive Context Aggregation (TRACE) to embed temporal and spatial information. The approach is as follows: from the visual features, relation candidates are represented as a hierarchical relation tree (HRTree); then, the TRACE module will capture temporal and spatial relationships to model the context with other frames; finally, a classification module will output the best inference.
%The challenge here is to create a model that intend only to past images as we are working with real-time stream. In fact, the method from \cite{Teng_Wang_Li_Wu_2021} attend to current methods of video SGG extend frame-level predictions to video-level by looking at past but also future images.

\textbf{Scene Graph Enrichment} Second, we enrich the graph with relevant commonsense data (Figure \ref{fig:sub1} bottom right). The challenge here is to bridge the gap between the ontology represented in the scene graph and the ontology represented in the commonsense knowledge graph. 
ConceptNet is a database of commonsense knowledge, it has been generated using multiple resources such as crowd-sourcing or expert-generated data. For each word in natural English language, ConceptNet relates other words or group of words using commonsense relationships such as "isUsedFor" or "is PartOf". For each relation, ConceptNet gives a set of connected entities. To select the most relevant, we pick the one with the highest confidence match with the data already present in the graph. For example, the relation "knife is used for cutting vegetables" will be selected over "knife is used for stabbing" if instances of vegetables (e.g. "tomato") are already present in the graph.
%In databases like ConceptNet there is a predefined set of 34 relations, whereas in a common scene graph the number of relations is not bounded. Moreover, in a scene graph, each node represents a label associated with bounding box coordinates \begin{math} \{x_1, y_2\} \end{math} (left-top corner and right-bottom corner, respectively). 
Thus, following \cite{Zareian_Karaman_Chang_2020}, we pair similar labeled nodes from both ontologies with a new edge. Then, all new edges are updated using successive message passing to propagate information across the graph. 
Grounding is applied by inferring related regions of the image to the corresponding enriched commonsense knowledge \cite{Zareian_Wang_You_Chang_2020}. This, for instance, will replace the word "vegetables" in the previous example by the one directly connected with the image, i.e. "tomato". The enrichment of commonsense knowledge is also used to interpret mis-used objects in a task, e.g. from the graph "PersonX is cutting a tomato with an axe" the commonsense knowledge from the word "axe" could be infrequently related to the word "tomato" and thus the task would be declared as \textit{unsafe}.  

\textbf{Sentiment Analysis on Graph} Third, sentiment analysis is performed given information from the graph (Figure \ref{fig:sub1} bottom left). We are evaluating words and their semantic connotation (e.g. the word "heat" is negatively connoted) using connotation lexicon such as \cite{Mohammad_Turney_2013}. Traditional approaches to build connotation lexicon rely on words prosody in texts, we want to extend this representation to visual features proximity using bounding boxes coordinates associated to every entity. For instance, spatial proximity between "negative" entities and the user in the image features will be highly weighted. We update the graph adding a \textit{sentiment value} \begin{math} S_n \in [-1;1] \end{math} to each node that will represent the potential risk of the entity for the human.

\textbf{Decision Making} Fourth, given the sentiment analysis, a pooling is performed with respect to the graph dependencies to retrieve a confidence value. If this value is above a pre-defined threshold, the task is declared as \textit{unsafe} and a decision of assistance is made. This threshold is dynamic and could be adjusted given contextual information such as the presence of a child in the scene.
%A complete overview of this architecture is shown in Figure \ref{fig:architecture}.

%\begin{figure}
%    \centering
%    \includegraphics[width=0.5\textwidth]{architecture.png}
%    \caption{Proposed architecture}
%    \label{fig:architecture}
%\end{figure}

%The threshold could be set as static or dynamic, changing with the type of task, e.g. if the task require the use of dangerous object such as knife it could be useful to increase the threshold.

%%ATOMIC relationships covers events causes (e.g. "Y implies Z"), agent intents (e.g. "personX wanted Y") as well as perception from others (e.g. "Because of Y, others think Z about personX") and many more.

\subsection{Command Generation}

Once the autonomous agent understands the immediate necessity of assistance, it needs to provide the appropriate help. To do so, the system needs to generate the best assistive action towards increasing the safety of the task, as in the example introduced in Section \ref{intro}.

%In the first case, we defined a task as understood if two or more successive sets of similar scene graphs are perceived. To do so, all scene graphs are save in the robot memory during runtime. The idea here is to step by step provide a natural language command that corresponds to the graph of the current event being performed. We will not reproduce human movements using for instance learning from demonstration methods, instead we build the required actions by setting high-level goals. Issues with learning from demonstration is that it is very dependent on the similarity between robot's and human's shape and the stability of the environment (i.e. no object should move between the demonstration and the reproduction). As we want our solution to be the most generic as possible we must disregard this approach. Similar to \cite{Zhu_Tremblay_Birchfield_Zhu_2021} our solution creates a high-level task planning as well as low-level motion generation. The task planning is retrieved from the scene graph, where only relevant triplets remains. Challenges here is to keep the temporal dependencies between sub-task movements. 

\begin{figure}
    \centering
    \includegraphics[width=0.5\textwidth]{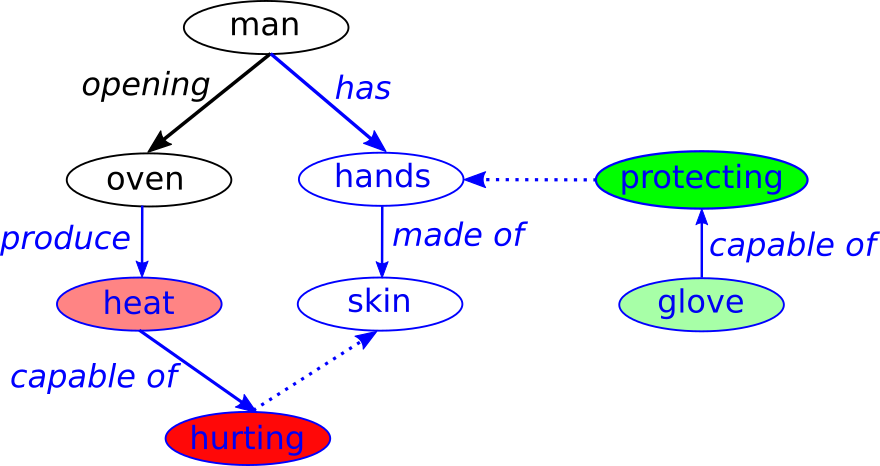}
    \caption{Solution retrieval via Scene Graph Completion.}
    \label{fig:SGC}
\end{figure}

We build what we call a "commonsense response", that means the most probable set of actions to perform for helping the human with the task. The goal here is to complete the weighted graph retrieved from the Sentiment Analysis with the related commonsense knowledge to stabilize the graph. We iterate through the graph using the same process as for Scene Graph Enrichment. The difference here is that for each iteration we also perform Sentiment Analysis and select only the positively weighted nodes. We compute the current sentiment value of the graph \begin{math}G\end{math} as follows:
\begin{equation}
  S(G) = \sum_{i=1}^{n} S_n
\end{equation}
where \begin{math}n\end{math} is number of nodes.
At the end of the process we obtain a graph similar to the one shown in Figure \ref{fig:SGC}, where the solution will be the highest positively weighted node that represents an object. This object could then be found and provide by the robot to the user. If no satisfying solution is found, one approach could be to warn the user by vocal utterances.

\subsection{Evaluation}

We will evaluate our Scene Graph Generation approach using the ActivityNet dataset \cite{caba2015activitynet}. This dataset contains 20k videos of 200 everyday life human activities and is used as a benchmark for most SGG approaches. Recently, the ActionGenome dataset was also introduced in \cite{ji2020action}. This dataset captures human daily life activities in 265k labeled frames. 

To assess user acceptance and perceived performance of the system, we will conduct a series of human experiments.
To do so, humans will engage in safe and potentially dangerous scenarios in which the system will detect risk of danger and define its remediating action to minimise detected danger. For practical and ethical reasons, humans will perform the safe task in a real environment and the potentially dangerous task in a Virtual Reality environment in which the task is simulated. We will use the Unified Theory of Acceptance and Use of Technology (UTAUT) \cite{venkatesh2007dead} and the Technology Acceptance Model 3 (TAM-3) \cite{venkatesh2008technology} to evaluate acceptance of the system in these scenarios. Additionally, we will supplement these measures with qualitative feedback about the performance and actions of the system. Human-Human interaction in similar scenarios will be used to evaluate the coherence of the assistive action of the system.

\section{Conclusion and future work} %% 1 page

The proposed work combines a traditional machine learning approach to generate an accurate model of the world with knowledge representation and reasoning. Our solution includes Scene Graph Generation as a high-level representation of the scene, commonsense knowledge enrichment combined with sentiment analysis to asses risk for the user and, finally, a graph completion method to retrieve relevant solutions.
As a limitation, our work does not consider other factors of the implicit need of help such as fatigue or stress. Furthermore, this proposal does not take into account the latent context of the captured sequence. For instance, in our running example it would be good to know if the oven is turned on or if it was turned on earlier.

As this proposal is still an on-going work, a number of challenges remain open, such as: the efficient fusion of scene and commonsense knowledge graphs; the sentiment analysis from scene graph and the generation of robot commands from graph entities. All these issues will be considered in future work, along with an investigation of the limitations, and confounding factors, in the automatic interpretation of the implicit need of help.

%In the first case, we defined a task as understood if two or more successive sets of similar scene graphs are perceived. To do so, all scene graphs are save in the robot memory during runtime. The idea here is to step by step provide a natural language command that corresponds to the graph of the current event being performed. We will not reproduce human movements using for instance learning from demonstration methods, instead we build the required actions by setting high-level goals. Issues with learning from demonstration is that it is very dependent on the similarity between robot's and human's shape and the stability of the environment (i.e. no object should move between the demonstration and the reproduction). As we want our solution to be the most generic as possible we must disregard this approach. Similar to \cite{Zhu_Tremblay_Birchfield_Zhu_2021} our solution creates a high-level task planning as well as low-level motion generation. The task planning is retrieved from the scene graph, where only relevant triplets remains. Challenges here is to keep the temporal dependencies between sub-task movements. 

\section{Acknowledgments} 
This publication was supported by
Brittany Region.

%% TOTAL: 8-9 pages

%%
%% Define the bibliography file to be used
\bibliography{biblio}

\begin{thebibliography}{30}
\expandafter\ifx\csname natexlab\endcsname\relax\def\natexlab#1{#1}\fi
\providecommand{\url}[1]{\texttt{#1}}
\providecommand{\href}[2]{#2}
\providecommand{\path}[1]{#1}
\providecommand{\DOIprefix}{doi:}
\providecommand{\ArXivprefix}{arXiv:}
\providecommand{\URLprefix}{URL: }
\providecommand{\Pubmedprefix}{pmid:}
\providecommand{\doi}[1]{\href{http://dx.doi.org/#1}{\path{#1}}}
\providecommand{\Pubmed}[1]{\href{pmid:#1}{\path{#1}}}
\providecommand{\bibinfo}[2]{#2}
\ifx\xfnm\relax \def\xfnm[#1]{\unskip,\space#1}\fi
%Type = Article
\bibitem[{Harnad(1990)}]{harnad1990symbol}
\bibinfo{author}{S.~Harnad},
\newblock \bibinfo{title}{The symbol grounding problem},
\newblock \bibinfo{journal}{Physica D: Nonlinear Phenomena}
  \bibinfo{volume}{42} (\bibinfo{year}{1990}) \bibinfo{pages}{335--346}.
%Type = Article
\bibitem[{Tada et~al.(2020)Tada, Hagiwara, Tanaka, and
  Taniguchi}]{tada2020robust}
\bibinfo{author}{Y.~Tada}, \bibinfo{author}{Y.~Hagiwara},
  \bibinfo{author}{H.~Tanaka}, \bibinfo{author}{T.~Taniguchi},
\newblock \bibinfo{title}{Robust understanding of robot-directed speech
  commands using sequence to sequence with noise injection},
\newblock \bibinfo{journal}{Frontiers in Robotics and AI} \bibinfo{volume}{6}
  (\bibinfo{year}{2020}) \bibinfo{pages}{144}.
%Type = Article
\bibitem[{Waldherr et~al.(2000)Waldherr, Romero, and
  Thrun}]{waldherr2000gesture}
\bibinfo{author}{S.~Waldherr}, \bibinfo{author}{R.~Romero},
  \bibinfo{author}{S.~Thrun},
\newblock \bibinfo{title}{A gesture based interface for human-robot
  interaction},
\newblock \bibinfo{journal}{Autonomous Robots} \bibinfo{volume}{9}
  (\bibinfo{year}{2000}) \bibinfo{pages}{151--173}.
%Type = Inproceedings
\bibitem[{Doisy(2012)}]{doisy2012sensorless}
\bibinfo{author}{G.~Doisy},
\newblock \bibinfo{title}{Sensorless collision detection and control by
  physical interaction for wheeled mobile robots},
\newblock in: \bibinfo{booktitle}{Proceedings of the seventh annual ACM/IEEE
  international conference on Human-Robot Interaction}, \bibinfo{year}{2012},
  pp. \bibinfo{pages}{121--122}.
%Type = Inproceedings
\bibitem[{Liu et~al.(2014)Liu, Zhang, and Li}]{Liu_Zhang_Li_2014}
\bibinfo{author}{R.~Liu}, \bibinfo{author}{X.~Zhang}, \bibinfo{author}{S.~Li},
\newblock \bibinfo{title}{Use context to understand user’s implicit
  intentions in activities of daily living},
\newblock in: \bibinfo{booktitle}{2014 IEEE International Conference on
  Mechatronics and Automation}, \bibinfo{year}{2014}, p.
  \bibinfo{pages}{1214–1219}.
%Type = Article
\bibitem[{Li and Zhang(2017)}]{Li_Zhang_2017}
\bibinfo{author}{S.~Li}, \bibinfo{author}{X.~Zhang},
\newblock \bibinfo{title}{Implicit intention communication in human–robot
  interaction through visual behavior studies},
\newblock \bibinfo{journal}{IEEE Transactions on Human-Machine Systems}
  \bibinfo{volume}{47} (\bibinfo{year}{2017}) \bibinfo{pages}{437–448}.
%Type = Article
\bibitem[{Kuipers(1984)}]{kuipers1984commonsense}
\bibinfo{author}{B.~Kuipers},
\newblock \bibinfo{title}{Commonsense reasoning about causality: deriving
  behavior from structure},
\newblock \bibinfo{journal}{Artificial intelligence} \bibinfo{volume}{24}
  (\bibinfo{year}{1984}) \bibinfo{pages}{169--203}.
%Type = Article
\bibitem[{Speer et~al.(2017)Speer, Chin, and Havasi}]{Speer_Chin_Havasi_2017}
\bibinfo{author}{R.~Speer}, \bibinfo{author}{J.~Chin},
  \bibinfo{author}{C.~Havasi},
\newblock \bibinfo{title}{Conceptnet 5.5: An open multilingual graph of general
  knowledge},
\newblock \bibinfo{journal}{Proceedings of the AAAI Conference on Artificial
  Intelligence} \bibinfo{volume}{31} (\bibinfo{year}{2017}).
%Type = Article
\bibitem[{Sap et~al.(2019)Sap, Bras, Allaway, Bhagavatula, Lourie, Rashkin,
  Roof, Smith, and
  Choi}]{Sap_Bras_Allaway_Bhagavatula_Lourie_Rashkin_Roof_Smith_Choi_2019}
\bibinfo{author}{M.~Sap}, \bibinfo{author}{R.~L. Bras},
  \bibinfo{author}{E.~Allaway}, \bibinfo{author}{C.~Bhagavatula},
  \bibinfo{author}{N.~Lourie}, \bibinfo{author}{H.~Rashkin},
  \bibinfo{author}{B.~Roof}, \bibinfo{author}{N.~A. Smith},
  \bibinfo{author}{Y.~Choi},
\newblock \bibinfo{title}{Atomic: An atlas of machine commonsense for if-then
  reasoning},
\newblock \bibinfo{journal}{Proceedings of the AAAI Conference on Artificial
  Intelligence} \bibinfo{volume}{33} (\bibinfo{year}{2019})
  \bibinfo{pages}{3027–3035}.
%Type = Article
\bibitem[{van Harmelen and Teije(2019)}]{van_Harmelen_Teije_2019}
\bibinfo{author}{F.~van Harmelen}, \bibinfo{author}{A.~t. Teije},
\newblock \bibinfo{title}{A boxology of design patterns for hybrid learning and
  reasoning systems},
\newblock \bibinfo{journal}{Journal of Web Engineering} \bibinfo{volume}{18}
  (\bibinfo{year}{2019}) \bibinfo{pages}{97–124}. \bibinfo{note}{ArXiv:
  1905.12389}.
%Type = Inproceedings
\bibitem[{Johnson et~al.(2015)Johnson, Krishna, Stark, Li, Shamma, Bernstein,
  and Fei-Fei}]{Johnson_Krishna_Stark_Li_Shamma_Bernstein_Fei-Fei_2015}
\bibinfo{author}{J.~Johnson}, \bibinfo{author}{R.~Krishna},
  \bibinfo{author}{M.~Stark}, \bibinfo{author}{L.-J. Li},
  \bibinfo{author}{D.~A. Shamma}, \bibinfo{author}{M.~S. Bernstein},
  \bibinfo{author}{L.~Fei-Fei},
\newblock \bibinfo{title}{Image retrieval using scene graphs},
\newblock in: \bibinfo{booktitle}{2015 IEEE Conference on Computer Vision and
  Pattern Recognition (CVPR)}, \bibinfo{publisher}{IEEE}, \bibinfo{year}{2015},
  p. \bibinfo{pages}{3668–3678}.
%Type = Inproceedings
\bibitem[{Feng et~al.(2011)Feng, Bose, and Choi}]{Feng_Bose_Choi_2011}
\bibinfo{author}{S.~Feng}, \bibinfo{author}{R.~Bose},
  \bibinfo{author}{Y.~Choi},
\newblock \bibinfo{title}{Learning general connotation of words using
  graph-based algorithms},
\newblock in: \bibinfo{booktitle}{Proceedings of the 2011 Conference on
  Empirical Methods in Natural Language Processing},
  \bibinfo{publisher}{Association for Computational Linguistics},
  \bibinfo{year}{2011}, p. \bibinfo{pages}{1092–1103}.
%Type = Inproceedings
\bibitem[{Li et~al.(2017)Li, Ouyang, Zhou, Wang, and
  Wang}]{Li_Ouyang_Zhou_Wang_Wang_2017}
\bibinfo{author}{Y.~Li}, \bibinfo{author}{W.~Ouyang},
  \bibinfo{author}{B.~Zhou}, \bibinfo{author}{K.~Wang},
  \bibinfo{author}{X.~Wang},
\newblock \bibinfo{title}{Scene graph generation from objects, phrases and
  region captions},
\newblock in: \bibinfo{booktitle}{2017 IEEE International Conference on
  Computer Vision (ICCV)}, \bibinfo{publisher}{IEEE}, \bibinfo{year}{2017}, p.
  \bibinfo{pages}{1270–1279}.
%Type = Article
\bibitem[{Ren et~al.(2017)Ren, He, Girshick, and
  Sun}]{Ren_He_Girshick_Sun_2017}
\bibinfo{author}{S.~Ren}, \bibinfo{author}{K.~He},
  \bibinfo{author}{R.~Girshick}, \bibinfo{author}{J.~Sun},
\newblock \bibinfo{title}{Faster r-cnn: Towards real-time object detection with
  region proposal networks},
\newblock \bibinfo{journal}{IEEE Transactions on Pattern Analysis and Machine
  Intelligence} \bibinfo{volume}{39} (\bibinfo{year}{2017})
  \bibinfo{pages}{1137–1149}.
%Type = Inproceedings
\bibitem[{Dai et~al.(2017)Dai, Zhang, and Lin}]{Dai_Zhang_Lin_2017}
\bibinfo{author}{B.~Dai}, \bibinfo{author}{Y.~Zhang}, \bibinfo{author}{D.~Lin},
\newblock \bibinfo{title}{Detecting visual relationships with deep relational
  networks},
\newblock in: \bibinfo{booktitle}{2017 IEEE Conference on Computer Vision and
  Pattern Recognition (CVPR)}, \bibinfo{publisher}{IEEE}, \bibinfo{year}{2017},
  p. \bibinfo{pages}{3298–3308}.
%Type = Inproceedings
\bibitem[{Zhang et~al.(2017)Zhang, Kyaw, Chang, and
  Chua}]{Zhang_Kyaw_Chang_Chua_2017}
\bibinfo{author}{H.~Zhang}, \bibinfo{author}{Z.~Kyaw}, \bibinfo{author}{S.-F.
  Chang}, \bibinfo{author}{T.-S. Chua},
\newblock \bibinfo{title}{Visual translation embedding network for visual
  relation detection},
\newblock in: \bibinfo{booktitle}{2017 IEEE Conference on Computer Vision and
  Pattern Recognition (CVPR)}, \bibinfo{year}{2017}, p.
  \bibinfo{pages}{3107–3115}.
%Type = Inproceedings
\bibitem[{Zellers et~al.(2018)Zellers, Yatskar, Thomson, and
  Choi}]{zellers2018neural}
\bibinfo{author}{R.~Zellers}, \bibinfo{author}{M.~Yatskar},
  \bibinfo{author}{S.~Thomson}, \bibinfo{author}{Y.~Choi},
\newblock \bibinfo{title}{Neural motifs: Scene graph parsing with global
  context},
\newblock in: \bibinfo{booktitle}{Proceedings of the IEEE Conference on
  Computer Vision and Pattern Recognition}, \bibinfo{year}{2018}, pp.
  \bibinfo{pages}{5831--5840}.
%Type = Inproceedings
\bibitem[{Yang et~al.(2018)Yang, Lu, Lee, Batra, and Parikh}]{yang2018graph}
\bibinfo{author}{J.~Yang}, \bibinfo{author}{J.~Lu}, \bibinfo{author}{S.~Lee},
  \bibinfo{author}{D.~Batra}, \bibinfo{author}{D.~Parikh},
\newblock \bibinfo{title}{Graph r-cnn for scene graph generation},
\newblock in: \bibinfo{booktitle}{Proceedings of the European conference on
  computer vision (ECCV)}, \bibinfo{year}{2018}, pp. \bibinfo{pages}{670--685}.
%Type = Article
\bibitem[{Wang et~al.(2020)Wang, Wei, Li, Zhang, and
  Huang}]{Wang_Wei_Li_Zhang_Huang_2020}
\bibinfo{author}{R.~Wang}, \bibinfo{author}{Z.~Wei}, \bibinfo{author}{P.~Li},
  \bibinfo{author}{Q.~Zhang}, \bibinfo{author}{X.~Huang},
\newblock \bibinfo{title}{Storytelling from an image stream using scene
  graphs},
\newblock \bibinfo{journal}{Proceedings of the AAAI Conference on Artificial
  Intelligence} \bibinfo{volume}{34} (\bibinfo{year}{2020})
  \bibinfo{pages}{9185–9192}.
%Type = Article
\bibitem[{Teng et~al.(2021)Teng, Wang, Li, and Wu}]{Teng_Wang_Li_Wu_2021}
\bibinfo{author}{Y.~Teng}, \bibinfo{author}{L.~Wang}, \bibinfo{author}{Z.~Li},
  \bibinfo{author}{G.~Wu},
\newblock \bibinfo{title}{Target adaptive context aggregation for video scene
  graph generation},
\newblock \bibinfo{journal}{Proceedings of the IEEE/CVF International
  Conference on Computer Vision}  (\bibinfo{year}{2021})
  \bibinfo{pages}{13688--13697}.
%Type = Inproceedings
\bibitem[{Li et~al.(2016)Li, Taheri, Tu, and Gimpel}]{Li_Taheri_Tu_Gimpel_2016}
\bibinfo{author}{X.~Li}, \bibinfo{author}{A.~Taheri}, \bibinfo{author}{L.~Tu},
  \bibinfo{author}{K.~Gimpel},
\newblock \bibinfo{title}{Commonsense knowledge base completion},
\newblock in: \bibinfo{booktitle}{Proceedings of the 54th Annual Meeting of the
  Association for Computational Linguistics (Volume 1: Long Papers)},
  \bibinfo{publisher}{Association for Computational Linguistics},
  \bibinfo{year}{2016}, p. \bibinfo{pages}{1445–1455}.
%Type = Article
\bibitem[{Bosselut et~al.(2019)Bosselut, Rashkin, Sap, Malaviya, Celikyilmaz,
  and Choi}]{Bosselut_Rashkin_Sap_Malaviya_Celikyilmaz_Choi_2019}
\bibinfo{author}{A.~Bosselut}, \bibinfo{author}{H.~Rashkin},
  \bibinfo{author}{M.~Sap}, \bibinfo{author}{C.~Malaviya},
  \bibinfo{author}{A.~Celikyilmaz}, \bibinfo{author}{Y.~Choi},
\newblock \bibinfo{title}{Comet: Commonsense transformers for automatic
  knowledge graph construction},
\newblock \bibinfo{journal}{arXiv:1906.05317 [cs]}  (\bibinfo{year}{2019}).
%Type = Inbook
\bibitem[{Zareian et~al.(2020)Zareian, Karaman, and
  Chang}]{Zareian_Karaman_Chang_2020}
\bibinfo{author}{A.~Zareian}, \bibinfo{author}{S.~Karaman},
  \bibinfo{author}{S.-F. Chang}, \bibinfo{title}{Bridging Knowledge Graphs to
  Generate Scene Graphs}, volume \bibinfo{volume}{12368} of
  \textit{\bibinfo{series}{Lecture Notes in Computer Science}},
  \bibinfo{publisher}{Springer International Publishing}, \bibinfo{year}{2020},
  p. \bibinfo{pages}{606–623}.
%Type = Inproceedings
\bibitem[{Feng et~al.(2013)Feng, Kang, Kuznetsova, and
  Choi}]{Feng_Kang_Kuznetsova_Choi_2013}
\bibinfo{author}{S.~Feng}, \bibinfo{author}{J.~S. Kang},
  \bibinfo{author}{P.~Kuznetsova}, \bibinfo{author}{Y.~Choi},
\newblock \bibinfo{title}{Connotation lexicon: A dash of sentiment beneath the
  surface meaning},
\newblock in: \bibinfo{booktitle}{Proceedings of the 51st Annual Meeting of the
  Association for Computational Linguistics (Volume 1: Long Papers)},
  \bibinfo{publisher}{Association for Computational Linguistics},
  \bibinfo{year}{2013}, p. \bibinfo{pages}{1774–1784}.
%Type = Article
\bibitem[{Zareian et~al.(2020)Zareian, Wang, You, and
  Chang}]{Zareian_Wang_You_Chang_2020}
\bibinfo{author}{A.~Zareian}, \bibinfo{author}{Z.~Wang},
  \bibinfo{author}{H.~You}, \bibinfo{author}{S.-F. Chang},
\newblock \bibinfo{title}{Learning visual commonsense for robust scene graph
  generation},
\newblock \bibinfo{journal}{arXiv:2006.09623 [cs]}  (\bibinfo{year}{2020}).
%Type = Article
\bibitem[{Mohammad and Turney(2013)}]{Mohammad_Turney_2013}
\bibinfo{author}{S.~M. Mohammad}, \bibinfo{author}{P.~D. Turney},
\newblock \bibinfo{title}{Crowdsourcing a word-emotion association lexicon},
\newblock \bibinfo{journal}{Computational Intelligence} \bibinfo{volume}{29}
  (\bibinfo{year}{2013}) \bibinfo{pages}{436–465}.
%Type = Inproceedings
\bibitem[{Fabian Caba~Heilbron and Niebles(2015)}]{caba2015activitynet}
\bibinfo{author}{B.~G. Fabian Caba~Heilbron, Victor~Escorcia},
  \bibinfo{author}{J.~C. Niebles},
\newblock \bibinfo{title}{Activitynet: A large-scale video benchmark for human
  activity understa\ nding},
\newblock in: \bibinfo{booktitle}{Proceedings of the IEEE Conference on
  Computer Vision and Pattern \ Recognition}, \bibinfo{year}{2015}, pp.
  \bibinfo{pages}{961--970}.
%Type = Inproceedings
\bibitem[{Ji et~al.(2020)Ji, Krishna, Fei-Fei, and Niebles}]{ji2020action}
\bibinfo{author}{J.~Ji}, \bibinfo{author}{R.~Krishna},
  \bibinfo{author}{L.~Fei-Fei}, \bibinfo{author}{J.~C. Niebles},
\newblock \bibinfo{title}{Action genome: Actions as compositions of
  spatio-temporal scene graphs},
\newblock in: \bibinfo{booktitle}{Proceedings of the IEEE/CVF Conference on
  Computer Vision and Pattern Recognition}, \bibinfo{year}{2020}, pp.
  \bibinfo{pages}{10236--10247}.
%Type = Article
\bibitem[{Venkatesh et~al.(2007)Venkatesh, Davis, and
  Morris}]{venkatesh2007dead}
\bibinfo{author}{V.~Venkatesh}, \bibinfo{author}{F.~Davis},
  \bibinfo{author}{M.~G. Morris},
\newblock \bibinfo{title}{Dead or alive? the development, trajectory and future
  of technology adoption research.},
\newblock \bibinfo{journal}{Journal of the association for information systems}
  \bibinfo{volume}{8} (\bibinfo{year}{2007}) \bibinfo{pages}{1}.
%Type = Article
\bibitem[{Venkatesh and Bala(2008)}]{venkatesh2008technology}
\bibinfo{author}{V.~Venkatesh}, \bibinfo{author}{H.~Bala},
\newblock \bibinfo{title}{Technology acceptance model 3 and a research agenda
  on interventions},
\newblock \bibinfo{journal}{Decision sciences} \bibinfo{volume}{39}
  (\bibinfo{year}{2008}) \bibinfo{pages}{273--315}.

\end{thebibliography}

\appendix

\end{document}